\documentclass[letterpaper]{article}

\usepackage{natbib,alifeconf}  
\usepackage{url,hyperref}
\usepackage{booktabs}

\usepackage{colortbl}
\usepackage{color}
\usepackage{multirow}
\usepackage{relsize}
\definecolor{lightgray}{gray}{0.9}

\usepackage{algorithm}
\usepackage{algpseudocode}
\usepackage{float}
\usepackage{amsmath}
\usepackage{cleveref}
\usepackage{comment}
\usepackage{nicematrix}

\usepackage{colortbl}
\usepackage{color}
\usepackage{multirow}
\usepackage{relsize}
\definecolor{lightgray}{gray}{0.9}

\algrenewcommand\algorithmicrequire{\textbf{Data:}}
\algrenewcommand\algorithmicensure{\textbf{Result:}}

\usepackage{graphicx}

\AtBeginDocument{%
  \providecommand\BibTeX{{%
    \normalfont B\kern-0.5em{\scshape i\kern-0.25em b}\kern-0.8em\TeX}}}

\usepackage{algorithm}
\usepackage{algpseudocode}
\usepackage{diagbox}
\usepackage{float}

\algrenewcommand\algorithmicrequire{\textbf{Data:}}
\algrenewcommand\algorithmicensure{\textbf{Result:}}

\begin{document}

\title{Untangling the Effects of Down-Sampling and Selection in Genetic Programming}




%

%

\author{
    Ryan Boldi$^{1}$,
    Ashley Bao$^{2}$,
    Martin Briesch$^3$,
    Thomas Helmuth$^4$, 
    Dominik Sobania$^3$, \\
    {\Large Lee Spector$^{2, 1}$ \and
    Alexander Lalejini$^5$} \\
    \mbox{}\\
    $^1$University of Massachusetts Amherst, Amherst MA, USA\\
    $^2$Amherst College, Amherst MA, USA \\
    $^3$Johannes Gutenberg University, Mainz, Germany \\
    $^4$Hamilton College, Clinton NY, USA \\
    $^5$Grand Valley State University, Allendale MI, USA \\
    rbahlousbold@umass.edu
}

\setlength\titlebox{2.5in}
%



\maketitle

\begin{abstract}
Genetic programming systems often use large training sets to evaluate the quality of candidate solutions for selection, which is often computationally expensive.
Down-sampling training sets has long been used to decrease the computational cost of evaluation in a wide range of application domains. 
More specifically, recent studies have shown that both random and informed down-sampling can substantially improve problem-solving success for GP systems that use the lexicase parent selection algorithm.
We test whether these down-sampling techniques can also improve problem-solving success in the context of three other commonly used selection methods, fitness-proportionate, tournament, implicit fitness sharing plus tournament selection, across six program synthesis GP problems.
We verified that down-sampling can significantly improve the problem-solving success for all three of these other selection schemes, demonstrating its general efficacy.
We discern that the selection pressure imposed by the selection scheme does not interact with the down-sampling method. However, we find that informed down-sampling can improve problem solving success significantly over random down-sampling when the selection scheme has a mechanism for diversity maintenance like lexicase or implicit fitness sharing. 
Overall, our results suggest that down-sampling should be considered more often when solving test-based problems, regardless of the selection scheme in use.

\end{abstract}


\section{Introduction}
Genetic programming (GP) applies the principles of Darwinian evolution to automatically synthesize programs instead of writing them by hand.  
GP systems are commonly used in the context of artificial life for both applied problem-solving~\citep{cava2021contemporary} and studying general evolutionary dynamics~\citep{dolson_digital_2021}, as evolved programs can express complex traits while still allowing researchers to fully disentangle the genetic mechanisms implementing those traits~\citep{lenski_evolutionary_2003,lalejini_adaptive_2021}. GP systems often use large training sets to evaluate the quality of candidate solutions (individuals). 
These training sets comprise examples of input and output pairs that describe the correct behavior of a program for a given problem. 
Each generation, individuals are evaluated on these pairs in order to determine whether or not they exhibit this desired behavior, such as returning the correct value for a program synthesis or regression problem. 
A parent selection algorithm then chooses the ``best'' individuals to contribute genetic material to the next generation. 

To thoroughly assess the quality of individuals in a population, most GP systems evaluate all individuals on every input-output example in the training set. 
This process can be computationally expensive when using large population sizes on large training sets or when individual evaluations are slow to compute. 
Reducing these computational costs can increase the scale at which we apply GP, allowing us to solve problems or conduct experiments that would otherwise not be possible. 
Down-sampling has been shown to be effective for reducing the per-generation cost of evaluating programs when using lexicase selection \citep{hernandez_random_2019,Ofria:2019:GPTP}.
Here, we show that these benefits apply to other selection methods, including tournament selection and fitness-proportionate selection.

Previous work demonstrated that using random down-sampling in the context of lexicase selection can substantially improve problem-solving success when the per-generation computational savings are reallocated to other aspects of evolutionary search, such as running for more generations \citep{hernandez_random_2019,Ofria:2019:GPTP,Helmuth2021benefits,schweim_effects_2022,geiger2023}.
However, naively constructing random down-samples has the drawback of leaving out potentially important training cases or over-representing redundant training cases, which can slow or even impede problem-solving success \citep{Hernandez2022,boldi_2022_environmental,Helmuth2021benefits,boldi_2023_static}.
Informed down-sampling (IDS) \citep{boldi_informed_2023} addresses this drawback by using runtime population statistics to construct down-samples with distinct, more informative training cases. 
Informed down-sampling was found to significantly improve success rates over random down-sampling for program synthesis runs using the PushGP system \citep{spector_push3_2005}.
In each of these previous studies, down-sampling is applied in the context of standard lexicase selection alone.
To our knowledge, these down-sampling techniques have yet to be rigorously evaluated in combination with other commonly used parent selection methods in GP, like tournament or fitness-proportionate selection, or other diversity-focused selection methods like implicit fitness sharing.



In this work, we expand on a previously published short-form communication~\citep{boldi2023problem} to investigate whether the benefits of random and informed down-sampling extend beyond lexicase selection. By doing this, we hope to motivate more artificial life practitioners to incorporate down-sampling into their evolutionary frameworks with a variety of selection schemes.
To do so, we compare problem-solving success when using different combinations of selection scheme and down-sampling method across six program synthesis benchmark problems~\citep{helmuth_general_2015,helmuth_psb2_2021}.
In addition, we test whether the level of selection pressure imposed by a selection scheme or whether the presence of diversity maintenance mechanisms influence the efficacy of random versus informed down-sampling. 

\section{Selection} 



The process of selection is a fundamental feature of evolutionary search. 
Parent selection algorithms steer evolving populations through a search space by determining which individuals should contribute genetic material to the next generation. 
Many selection algorithms have been developed, each targeting different problem domains and search space topologies (e.g., \citep{holland_adaptation_1992,Brindle1980,ross_effects_1999,assessment_spector_2012,Helmuth_solving_2015}). 
In this subsection, we overview the four selection strategies studied in this paper: fitness-proportionate, tournament, implicit fitness sharing and lexicase selection. 
The implementation details and specific parameters of each of these strategies are then discussed in their respective subsection.

We acknowledge that there are many different selection schemes that were not included in this study, including those based on rankings \citep{blickle1996comparison}, elitism, quality diversity \citep{Mouret2015IlluminatingSS}, or other techniques. However, we believe that this set of selection schemes captures a significant portion of what GP practitioners use. 

\subsection{Fitness-Proportionate Selection}

Fitness-proportionate selection (FPS) is one of the earliest proposed selection strategies in evolutionary computation \citep{holland_adaptation_1992}. 
Fitness-proportionate selection assigns each parent a selection probability based on its aggregate fitness relative to that of the other population members. 
Although individuals with higher fitness have a higher chance of being selected, those with lower fitness can still be selected. The probability $p_i$ that an individual $i$ is selected is $$p_i = \dfrac{f_i}{\mathlarger{\sum}_{j=1}^N\, f_j}$$ where $f_i$ is the $i^\text{th}$ individual's fitness, and $N$ is the population size. 
Since our genetic programming system evaluates individuals with \emph{errors} instead of fitness values, we compute the fitness of the individual as $\frac{1}{1+e_i}$ where $e_i$ is the aggregate error the individual achieved on the training set.

On its own, fitness-proportionate selection can impose low selection pressure on a population relative to other commonly used selection algorithms, such as tournament selection~\citep{blickle1996comparison,Zhong_tourn_2005}.
Fitness-proportionate selection is also simple to implement and computationally efficient with a time complexity of $\mathcal{O}(N)$.
As such, fitness-proportionate selection is still commonly used for evolutionary search \citep{Dang2019RuntimeAO,Arabas2020PopulationDO}, often as one component of more sophisticated selection procedures \citep{YanFPS}. 

\subsection{Tournament Selection}
Tournament selection requires there to be a comparison metric between individuals. This can either be represented as a total ordering (ranking) of the individuals, or an assignment of a fitness value for each individual.
To select a single individual with tournament selection, $t$ individuals are chosen from the population at random.
Then, the individual with the best fitness (or lowest error) ``wins'' the tournament and is selected as a parent. 
Tournament size ($t$) controls the strength of selection; larger tournament sizes impose stronger selection, and smaller tournament sizes impose weaker selection.
In this work, we use a variety of tournament sizes.

Tournament selection has been found to be more stable \citep{Butz2003TournamentSS} than fitness-proportionate selection, as it is not affected by fitness scaling \citep{goldberg_comparative_1991}. 
Tournament selection also has a time complexity of $\mathcal{O}(N)$, which makes it attractive when using large population sizes \citep{goldberg_comparative_1991}. 
Due to its simplicity and efficiency, it is widely used as the standard selection strategy for evolutionary computation \citep{fang_review_2010}. 

\subsection{Implicit Fitness Sharing}
Implicit fitness sharing (IFS) changes how errors are aggregated to incorporate an estimate of the difficulty of a training example \citep{whitley93IFS,ifsGPMcKay00}. This technique has been used as a diversity preservation method as individuals that solve hard or rare test cases have a higher fitness than otherwise. IFS was later adapted to enable it to be used in cases where errors are non binary \citep{Krawiec13IFS}, which is the framework we use in this work. Implicit fitness sharing is not a selection scheme on its own, but a fitness augmentation scheme that can be applied before using a different selection scheme, such as tournament or fitness-proportionate selection to make the selection. 

The idea for implicit fitness sharing is that individual test case error is weighted with respect to how the rest of the population performs on that test case. Specifically, the corrected fitness for each individual is given by the formula

$$F_{NBIFS}(i) = \sum_{t\in T}\dfrac{f(i, t)}{\sum_{i'\in P} f(i', t)}$$

Where $T$ is the set of training cases, $f(i, t) \in [0, 1]$ is the raw fitness of an individual $i$ on training case $t$ (higher is better). We calculate this value from the errors of the individual in the same manner as we did for fitness-proportionate Selection. After this re-weighting scheme, the population is selected from using tournament selection. In this work, we use a tournament size of $t=30$ for the IFS experiments, as this tournament size was empirically determined to perform the best from our plain tournament selection runs. When we refer to IFS as a selection scheme, we are referring to IFS with tournament selection with $t=30$.

\subsection{Lexicase Selection}

Unlike fitness-proportionate and tournament selection, lexicase selection does not aggregate error values across training cases to choose parents. 
Instead, lexicase considers each training case individually. 
To select a single parent, lexicase selection first shuffles the set of training cases into a random order, and all individuals in the population are included in a pool eligible for selection.
Each training case is then applied in sequence (in the shuffled order) to filter down the pool of eligible candidates. At each step, the next pool of eligible individuals is set to include only individuals with elite performance on the current training case.  
This filtering continues until one individual remains in the eligible pool to be selected or until all training cases have been exhausted, where one of the remaining eligible individuals is then selected at random.
Because each parent selection event uses a random permutation of training cases, lexicase selection prioritizes high performance on different sets of training cases across parent selection events, improving its capacity for diversity maintenance \citep{helmuth_effects_2016,dolson_ecological_2018}.

Lexicase selection was initially designed for multi-modal test-based program synthesis problems \citep{assessment_spector_2012,Helmuth_solving_2015} and has frequently been found to outperform other selection methods in this domain~\citep{helmuth_applying_2022,sobania2022comprehensive}.
Lexicase selection has also been shown to be effective in domains beyond GP, including evolutionary robotics \citep{moore2017lex,stanton_lexicase_2022}, deep learning \citep{Ding2022optimizing}, genetic algorithms \citep{metevier_lexicase_2019}, learning classifier systems \citep{Aenugu2019} and even in the directed evolution of microbes \citep{lalejini_artificial_2022}. 
Lexicase selection's success is often attributed to its capacity to preserve diversity \citep{helmuth_effects_2016,dolson_ecological_2018} and maintain specialists \citep{helmuth_importance_2020}. 
However, the worst case time complexity for lexicase selection is $\mathcal{O}(N*C)$ where $N$ is the population size, and $C$ is the number of training cases. 
In practice, this number is often closer to $\mathcal{O}(N+C)$ when population diversity is high \citep{helmuth_population_2022}, and there are strategies that can be used to reduce this even further \citep{ding2022scale, Lalejini2023PhylogenyinformedFE}. Despite this larger time complexity, in practice, the computational cost of genetic programming is more often dominated by program evaluation instead of selection. 
In this work, we focus on training set down-sampling, a strategy that can be used to reduce how expensive the evaluation step of evolutionary runs are by reducing the effective size of the training set. 

\section{Down-sampling}

In test-based program synthesis, candidate solutions are evaluated on a set of training cases in order to assess quality or correctness. Down-sampling techniques reduce the total number of training cases used for assessing candidate solution quality, which in turn, reduces the total number of program evaluations needed each generation. 
Down-sampling has been studied in evolutionary computation as a means to reduce computational loads \citep{langdon_minimising_2011,ross_effects_1999} and reduce overfitting \citep{goncalves_random_2012,liu_reducing_2004,goncalves:2013:EuroGP}. 

\emph{Historical subset selection} is one simple down-sampling method that maintains a single static subset of the training cases for an entire evolutionary run \citep{ga94aGathercole}.
In contrast, \emph{random subset selection} \citep{ga94aGathercole} randomly chooses to include or not include each training case each generation, resulting in different down-sample sizes from generation to generation. \emph{Stochastic subset sampling} \citep{nordin_-line_1997,lasarczyk_dynamic_2004} chooses a new fixed-size down-sample each generation. 
More sophisticated methods of down-sampling have also been developed.  
\emph{Dynamic subset selection} creates down-samples that are biased toward including harder cases and cases not seen for many generations \citep{ga94aGathercole}. Other work introduces a topology-based selection that takes problem structure into account by selecting cases that individuals perform differently in a problem domain \citep{lasarczyk_dynamic_2004}. Topology-based selection is very similar to the later proposed informed down-sampling \citep{boldi_informed_2023}; although the latter has only been tested on program synthesis problems, and the former only on symbolic regression and classification. 
In this work, we compare the performance of informed down-sampling to that of a method similar to the earlier proposed \emph{stochastic subset sampling}, also known as random down-sampling \citep{hernandez_random_2019}, and the standard non-down-sampled selection strategy.


Given the demonstrated value of down-sampling for evolutionary search, it is important to understand how different down-sampling methods interact with different selection algorithms to benefit (or hinder) problem-solving success.
Here, we focus on two down-sampling techniques that have been demonstrated to be effective in combination with lexicase selection for GP: random down-sampling and informed down-sampling, each described in detail below. 
We ask whether the benefits of these down-sampling methods might extend to other commonly used selection procedures in GP.

\begin{algorithm*}
\caption{Informed Down-Sampling. Adapted from \citep{boldi_informed_2023}.}
\label{alg:IDSTotal}
\begin{algorithmic}[1]
\Require
\Statex $\mathcal{P}:$ population,\,\,\, $\textbf{cases}$:  set of all training cases,
\Statex$\mathcal{S}: $ selection scheme, \Comment{$\mathcal{S}$ picks a new pop. given an old pop. and a set of cases}
\Statex$k: $ scheduled case distance computation parameter, 
\Statex$\rho: $ parent sampling rate,  \Comment{$\rho, k$ are parameters to reduce the distance computation cost}
\Statex $\mathcal{G}:$  current generation counter,
\Statex $\mathcal{D}: $ case distance matrix. \Comment{all distances are initialized to be maximally far}
\Ensure A list of selected parents
\If{$\mathcal{G} \% k == 0$}
\State{$\hat{\mathcal{P}} \gets$ sample $\rho{\times}|\mathcal{P}|$ parents from $\mathcal{P}$}
\State{evaluate $\hat{\mathcal{P}}$ on $\textbf{cases}$} \Comment{parent sample, purely used for distance calculations}
\State{calculate $\mathcal{D}$ from solve vectors from solutions in $\hat{\mathcal{P}}$ on $\textbf{cases}$}
\EndIf
\State{$\textbf{ds} \gets $ create down-sample using farthest first traversal} \Comment{picks cases that are of high distance to each other in a greedy fashion}
\State{$\mathcal{P} \gets$ select $|\mathcal{P}|$ new parents using $\mathcal{S}$ from $\mathcal{P}$ using $\textbf{ds}$ as cases}  \Comment{selecting new population}
\State \bf{return} $\mathcal{P}$
\end{algorithmic}
\end{algorithm*}


\subsection{Random Down-sampling}

Random down-sampling (in this context) constructs a random fixed-size subset of the training set each generation.
This smaller subset of training cases is then used to evaluate the quality of the population for selection in the current generation.
By reducing the number of training cases used to evaluate programs each generation, random down-sampling reduces the per-generation computational costs of population evaluation and parent selection. 
Previous work demonstrated that reallocating these computational savings to other aspects of evolutionary search can lead to substantial improvements in problem-solving success in the context of lexicase selection~\citep{hernandez_random_2019,Ofria:2019:GPTP,Helmuth2021benefits,schweim_effects_2022,geiger2023}. 

However, random down-sampling results in less thorough program evaluations, which can lead to misleading assessments of program quality. 
For example, a random down-sample might omit important training cases (e.g., cases that test input edge cases), as the down-sample is created randomly with no consideration for the program behavior that each training case might be assessing.  
Prior work explored the extent to which random down-sampling resulted in discontinuities between training sets used to select successive generations \citep{boldi_2022_environmental}. They found that the commonality of synonymous training cases usually prevented discontinuities; that is, most training sets contain many training cases that measure the same behavior are thus passed by the same groups of individuals. 
In these circumstances, down-samples are less likely to entirely omit an entire class of training cases. 

\subsection{Informed Down-sampling}

Informed down-sampling addresses random down-sampling's drawback of potentially omitting informative training cases by minimizing the number of synonymous training cases included in the down-sample \citep{boldi_informed_2023}. 
To estimate differences among training cases, informed down-sampling fully evaluates a random subset of the population on the complete set of training cases. If an individual has 0 error on a training case, we refer to this individual as ``solving" that training case. To select an individual, we evaluate them on a subset of the training cases. Two training cases solved by the exact same subset of the population are less informative than having two training cases that are solved by very different sub-populations. This is because if both cases are solved by the same set of the population, they exert very similar selective pressure to if there was only one of those cases. Following prior work, we call these cases ``synonymous" due to the fact that they measure same behavior in the context of the current population. With informed down-sampling, down-samples are constructed to include cases as far from being synonymous with each other as possible.

Algorithm~\ref{alg:IDSTotal} specifies the full informed down-sampling procedure. 
We modified the algorithm from \citep{boldi_informed_2023} to allow for an arbitrary selection scheme, $\mathcal{S}$.
To construct a down-sample, we use a random subset of the population to estimate the ``distance'' between all pairs of training cases. 
The distance between two training cases is the Hamming distance between their ``solve vectors'', which are vectors of binary value that specify which individuals in the population subset (or ``parent sample") solved the training case.
We specify the size of the population subset (and thus the maximum distance between cases) using the $\rho$ parameter. 
With $\rho = 0.01$, we include 1\% of the population in the population subset used for calculating each training case's solve vector. 
We used $\rho = 0.01$ for all experiments in this work. 
Next, a single training case is added to the down-sample at random, and training cases are added in sequence such that each additional training case is maximally far away from the current down-sample through a process known as farthest first traversal \citep{Hochbaum1985ABP}.


Whilst what we outlined above is one specific specification of informed (or non-random) down-sampling, there are several other methods that can fall under this umbrella term. We believe the above outlined informed down-sampling sufficiently captures the main ideas from semantic aware down-sampling. It does not use information regarding inputs and outputs to form a down-sample. Instead, it uses population statistics, which exist regardless of the problem being solved, which makes it applicable in all problem domains without problem specific augmentations needed. For this reason, we think it is the most general approach to semantic aware down-sampling and we include it as the representative algorithm in this work.

\section{Methods}
In this work, we study random and informed down-sampling in the context of four selection schemes commonly used in GP: tournament selection, fitness-proportionate selection, implicit fitness sharing and lexicase selection.

Our first series of experiments analyzes the performance of each specific selection and down-sampling combination. 
The second set of experiments focuses on determining how varying the strength of selection pressure influences the efficacy of down-sampling.  For each of these experiments, we compare problem-solving success on six program synthesis problems as detailed in the following section.

\subsection{Program Synthesis Problems} 
The goal of program synthesis problems is to achieve zero error on each of a set of training cases, where each training case encodes what the program should output given a certain input. For this work, we chose six program synthesis problems from the first and second program synthesis benchmark suites \citep{helmuth_general_2015,helmuth_psb2_2021,helmuth_applying_2022}. 
These problems have been explored in previous work on informed down-sampling \citep{boldi_informed_2023} and are therefore a good basis for this investigation.  
We included problems where informed down-sampling has been shown to improve problem-solving success (Count Odds, Fizz Buzz and Scrabble Score), reduce problem-solving success (Small or Large), and have no significant effect on problem-solving success (Fuel Cost). We also include a new problem that has not been investigated in prior work on informed down-sampling (Middle Character).

The specific parameters used for our program synthesis experiments can be found in Table~\ref{tab:ProgSynthParams}. 
We performed 50 evolutionary runs for each program synthesis problem configuration, each with a population size of 1000. 
Each of these runs were performed at 5\% down-sampling, meaning $r=0.05$.
Since we have 200 training cases in the entire training set, each individual is evaluated on 10 of the cases every generation. 
We chose $r=0.05$ based on previous work \citep{boldi_informed_2023}.
The down-sampling strategy being used determines which cases are selected out of the 200 to make up the down-sample.

\begin{table}[t]
    \centering
    \caption{System parameters used for the program synthesis runs. We seperate the parameters by general GP parameters as well as parameters used specifically when down-sampling is enabled (DS).}
    \setlength{\tabcolsep}{17pt}
    \resizebox{0.9\columnwidth}{!}{
    \begin{tabular}{|lr|}
    \hline
    \textbf{GP Parameter}     & \textbf{Value}\\
    \hline
    runs per problem & 50 \\
    population size & 1000 \\
    initial training set size & 200 \\
    testing set size & 1000 \\
    maximum program executions & 60,000,000 \\
    variation operator & UMAD \\
    \hline
     \textbf{DS Parameter} & \textbf{Value}\\
     \hline
     down-sample rate $r$  & 0.05\\
     parent sample rate $\rho$ & 0.01 \\
     generational interval $k$ & 100 \\
    \hline
    \end{tabular}
    }
    \label{tab:ProgSynthParams}
\end{table}

\begin{table*}[t!]
\centering
\caption{The effect that varying the selection schemes has on problem solving success when in conjunction with down-sampling. We report 
the number of generalizing solutions out of 50 program synthesis runs achieved by PushGP on the test set. In \textbf{bold} font are the down-sampling runs that perform significantly better than the respective runs with no down-sampling. Significant differences (according to a pairwise Chi-Squared significance test) between informed and random down-sampling are denoted by an asterisk (*).
}\label{tab:selection_scheme_results}
\setlength{\tabcolsep}{5pt}
\renewcommand{\arraystretch}{1.5} 
\resizebox{0.9\textwidth}{!}{\begin{tabular}{l|l|ccc|ccc|ccc|ccc|}
& Selection Scheme & \multicolumn{3}{c}{Fitness Prop.} & \multicolumn{3}{c}{Tourn. ($t=30$)} & \multicolumn{3}{c}{IFS ($t=30$)} & \multicolumn{3}{c}{Lexicase} \\
\cline{2-14}
& Down-sample Type & No & Rnd & IDS & No & Rnd & IDS  & No & Rnd & IDS & No & Rnd & IDS \\
\hline
& Count Odds \,\,\,\,\,\, & 0 & 1 & 0 & 1 & \textbf{31} & \textbf{34} & 2 & \textbf{22} & \textbf{42}* & 11 & 10 & \textbf{49}* \\
\rowcolor{lightgray}
\cellcolor{white} & Fizz Buzz & 0 & 5 & 5 & 0 & 4 & \textbf{7} & 0 & 2 & 4 & 5 & \textbf{32} & \textbf{45}* \\ 
& Fuel Cost & 0 & \textbf{19} & \textbf{14} & 0 & \textbf{28} & \textbf{37} & 0 & \textbf{26} & \textbf{40}* & 22 & \textbf{40} & \textbf{41} \\
\rowcolor{lightgray}
\cellcolor{white} & Small or Large & 0 & 0 & 0 & 11 & \textbf{37} & \textbf{41} & 13 & 18 & \textbf{47}* & 16 & \textbf{42} & \textbf{38} \\
\rowcolor{white}
\cellcolor{white} & Middle Character & 0 & 0 & 0 & 4 & \textbf{17} & 10 & 1 & \textbf{15} & \textbf{16} & 16 & \textbf{30} & 26 \\
\rowcolor{lightgray}
\multirow{-6}{*}{\rotatebox{90}{Problem}} \cellcolor{white} & Scrabble Score & 0 & 0 & 0 & 0 & 1 & 0 & 0 & 0 & 0 & 1 & 0 & 2 \\
\hline
\end{tabular}}
\end{table*}

To make the comparisons fair, we ensure that all methods (regardless of down-sampling) use the same number of program executions. 
We limited all runs to 60,000,000 program executions, which is equivalent to running a full GP run (with no down-sampling) for 300 generations.

\subsection{Experimental Design }
We used the PushGP framework to conduct our experiments. 
PushGP is a genetic programming framework for evolving Push programs. 
The Push programming language uses a set of typed memory stacks to allow programs to handle different data types (e.g., strings, numbers, etc.) and includes a Turing complete instruction set that supports basic computations as well as complex control flow, such as looping and conditional execution \citep{spector2:2001:gecco,spector_genetic_2002,spector_push3_2005}. For these experiments, we use the same instruction sets as those used by \citet{boldi_informed_2023}, and 
used the propeller\footnote{https://github.com/lspector/propeller} implementation of PushGP.

For each configuration, we report the number of generalizing runs, which is the number of runs that produce a program that passes all test cases in the held out testing set. Since we are not evaluating the individuals on the entire training set when we down-sample, checking if an individual passes the \emph{entire} training set happens when an individual passes all the cases in the down-sample. If an individual passes the entire training set, the evolutionary run ends. This individual is then evaluated on the held out testing set. If this individual passes the testing set, the run is marked as a generalizing run. If the individual passes the down-sample, but not the entire \emph{training} set, the evolutionary run continues. Contrary to previous work on down-sampling with lexicase selection \citep{hernandez_random_2019,Helmuth2021benefits,boldi_informed_2023}, the extra program executions required to verify if an individual passes the entire training set are added to our program execution tally used to limit our runs.

\subsection{Diversity preservation properties}


\begin{table*}[t!]
\centering
\caption{The effect that varying the selection pressure (tournament size) has on problem solving success when in conjunction with down-sampling. We report 
the number of generalizing solutions (successes) out of 50 program synthesis runs achieved by PushGP on the test set.
}\label{tab:tsize_results}
\setlength{\tabcolsep}{5pt}
\renewcommand{\arraystretch}{1.5} 
\resizebox{0.9\textwidth}{!}{\begin{tabular}{l|l|ccc|ccc|ccc|ccc|}
& Tournament Size & \multicolumn{3}{c}{$t=2$} & \multicolumn{3}{c}{$t=5$}  & \multicolumn{3}{c}{$t=10$} & \multicolumn{3}{c}{$t=30$} \\
\cline{2-14}
& Down-sample Type & No & Rnd & IDS & No & Rnd & IDS  & No & Rnd & IDS & No & Rnd & IDS \\
\hline
& Count Odds & 1 & 0 & 0 & 0 & \textbf{7} & \textbf{10} & 0 & \textbf{26} & \textbf{33} & 1 & \textbf{31} & \textbf{34} \\
\rowcolor{lightgray}
\cellcolor{white} & Fizz Buzz & 0 & 0 & 0 & 0 & 0 & 0 & 0 & 0 & 1 & 0 & 4 & \textbf{7} \\
& Fuel Cost & 0 & 0 & 0 & 0 & \textbf{14} & \textbf{11} & 1 & \textbf{28} & \textbf{25} & 0 & \textbf{28} & \textbf{37} \\
\rowcolor{lightgray}
\cellcolor{white} & Small or Large & 0 & 2 & 2 & 7 & 4 & \textbf{21}* & 13 & 18 & \textbf{47}* & 11 & \textbf{37} & \textbf{41} \\ 
& Middle Character & 0 & 1 & 1 & 0 & 0 & 3 & 1 & \textbf{13} & \textbf{10} & 4 & \textbf{17} & 10 \\
\rowcolor{lightgray}
\multirow{-6}{*}{\rotatebox{90}{Problem}} \cellcolor{white} & Scrabble Score & 0 & 0 & 0 & 0 & 0 & 0 & 0 & 1 & 0 & 0 & 1 & 0 \\
\hline
\end{tabular}}
\end{table*}

A specific quality of selection schemes that could have an effect when used in conjunction with down-sampling is the diversity preservation qualities of the scheme. 

Diversity maintenance is the capacity for a selection method to select a behaviorally diverse population of individuals. Estimating the behavior of an individual (so that you can maintain diversity) might be less accurate with the addition of down-sampling as a system has access to less information about the individual's behavior.

We investigate whether the diversity maintenance properties of a selection scheme affects its compatibility with down-sampling. To do this, we evaluate the problem-solving performance of selection schemes that use a variety of diversity preservation qualities. Specifically, we compare two selection schemes with no diversity preservation (FPS and Tournament) and two that do (Lexicase and IFS).

\subsection{Selection Pressure}
We investigate how selection pressure interacts with down-sampling by comparing problem-solving success of tournament selection at a range of tournament sizes.  

Selection pressure is the standard that is required to be met in order for an individual to be selected on average. With a high selection pressure, an individual needs to be better than more of its peers to have the same chance at selection. It is important to study whether the selection pressure affects the efficacy of down-sampling. This is because if the fitness evaluation is noisy due to an unbalanced sample, high selective pressure might result in worse selections being made. In other words, being highly selective with respect to an non-representative sample could result in selecting individuals that are globally suboptimal.

In order to vary the selection pressure, we vary the size of the tournament. We test tournament sizes of $t=2, 5, 10, 30$ in combination with the various types of down-sampling.

\section{Results and Discussion}

Table~\ref{tab:selection_scheme_results} shows problem-solving successes for the six program synthesis problems studied at a variety of selection scheme and down-sampling combinations. 
A run is considered to be successful if a perfect solution evolves (i.e., a program that solves all training and unseen testing cases).
Consistent with previous work with lexicase selection~\citep{Helmuth2021benefits,helmuth_applying_2022,Hernandez2022,boldi_informed_2023}, not all program synthesis problems benefited from down-sampling.
However, we found no instances where configurations without down-sampling significantly outperformed configurations with down-sampling enabled.
In fact, when using lexicase selection and tournament selection, problem-solving success was significantly improved for all problems but one by at least one of the down-sampling method. When using implicit fitness sharing, down-sampling significantly improved problem-solving success for all but two problems (Scrabble Score and Fizz Buzz).
Overall, fitness-proportionate selection benefited the least from the addition of down-sampling, as problem-solving success was significantly better for only one out of the six problems.
We also found some examples where fitness-proportionate and tournament selection failed to find \textit{any} solutions unless we used down-sampling. 


We detected a significant difference in problem-solving success between informed and random down-sampling in five instances (across all problems and configurations). 
With implicit fitness sharing, we found significant differences between informed down-sampling and random down-sampling on the Count Odds, Fuel Cost and Small or Large problems. For lexicase selection, we found a significant difference on the Count Odds and Fizz Buzz problems. 
In each of these instances, informed down-sampling outperformed random down-sampling. We found no significant differences between the two down-sampling techniques on any problems when using fitness-proportionate or Tournament selection with $t=30$.


To better understand what might distinguish selection schemes for which downsampling helps from those for which it doesn't, we  consider the role that the overall selection pressure of a selection scheme might play. One might expect, for example, that when using an informed down-sample it would be beneficial to maintain high selection pressure with respect to those well-chosen cases, whereas high selection pressure could result in the loss of test coverage on the cases not in the down-sample when using random down-sampling. To cast some light on this and related possibilities, we conducted experiments using tournament selection, for which selection pressure is easily adjusted.

The results for a variety of different configurations of tournament selection can be found in Table~\ref{tab:tsize_results}. We find that varying the selection pressure imposed by tournament selection does have a significant effect on problem solving success. Over these configurations, we find multiple places where a down-sampling technique outperforms the non downsampled version, yet we only find one problem (Small or Large) where informed down-sampling outperforms random down-sampling significantly. For this reason, it seems as though selective pressure does not significantly affect the comparative performance between random and informed down-sampling. Given this, we can be reasonably certain that down-sampling will improve the performance of their systems, regardless of the selection pressure enacted by the selection scheme chosen.


Overall, our results indicate that down-sampling is often beneficial or neutral for problem-solving success. 
We did not find compelling evidence that down-sampling \textit{impeded} success in any of our experiments.
Though, we do note that others have found down-sampling to impede problem-solving success when there are strong trade-offs between training cases (e.g., low error on one excludes low error on another) or when a training set lacks some redundancy \citep{Hernandez2022}.


We found that informed down-sampling was most consistently beneficial in the context of lexicase selection and implicit fitness sharing, as problem-solving success was improved by at least one down-sampling method across all problems for both of these selection schemes.
We hypothesize that this is due to these schemes' ability to maintain diverse populations. 
Fitness-proportionate and tournament selection are known to be susceptible to premature convergence~\citep{hornby_alps_2006,hernandez_diagnostics_2022}, while both lexicase selection and implicit fitness sharing are capable of maintaining both phenotypic and phylogenetic diversity \citep{dolson_ecological_2018,riolo_lexicase_2016,shahbandegan_untangling_2022,hernandez_what_2022}. 
Given this, we hypothesize that lexicase selection and implicit fitness sharing benefit more from the increased number of generations afforded by down-sampling, and the presence of distinct and diverse training cases afforded by the ``informed-ness", than tournament or fitness-proportionate selection. This also relates to the results of a different study, that showed that informed down-sampling maintains higher test coverage from successive selections than random down-sampling \citep{boldi_2023_static}.
That is, if a population evolving under fitness-proportionate and tournament selection has converged to a local fitness optimum, that population may not benefit from extra generations of evolution.
In contrast, a more diverse population evolving under lexicase selection or IFS may benefit substantially from running for an increased number of generations when they are evaluated on a diverse set of training cases. Recent work also hints at the merit of increasing a population size instead of increasing the number of generations, potentially shrinking the gap between random and informed down-sampling \citep{Briesch2023OnTT}.

\section{Conclusion}


In this work, we extended previous studies that evaluated the efficacy of random and informed down-sampling in the context of lexicase selection \citep{hernandez_random_2019,Helmuth2021benefits,boldi_informed_2023}.

Here, we show that the problem-solving benefits of both random and informed down-sampling generalize to other selection schemes, including fitness-proportionate selection, tournament selection and other diversity preserving selection schemes like implicit fitness sharing (IFS).
This result suggests that evolutionary computing practitioners should experiment with different forms of down-sampling in combination with their preferred selection methods, as it can be used to improve problem-solving success by reallocating per-generation computational savings to running a deeper evolutionary search.

Previous studies have shown that the benefits of down-sampling stem from reallocating the computational savings to running an evolutionary search for more generations or evaluating more individuals~\citep{Helmuth2021benefits,hernandez_random_2019,Ofria:2019:GPTP}.
We hypothesize that this explanation holds across each of the selection schemes that we tested in this work. 
We did, however, find that different selection schemes benefited more or less from the addition of down-sampling: fitness-proportionate selection seemed to benefit the least, while lexicase, implicit fitness sharing, and even tournament selection with $t=30$ benefited from down-sampling on five of the six problems. 

We also detected that some selection schemes benefit more or less from the inclusion of informed down-sampling. Specifically, we found that for both lexicase selection and implicit fitness sharing multiple configurations using informed down-sampling significantly improved problem solving success over random down-sampling by including more unique and distinct cases in the down-samples.
We hypothesize that populations evolving under lexicase selection or IFS are more diverse and therefore benefit the most from the extra generations of informative cases that are afforded by informed down-sampling. 

To test the impact of selection pressure on down-sampling, we adjusted the tournament size for tournament selection and observed if it affected the performance of the two down-sampling techniques. Changing the tournament size did not consistently influence the relative performance between the two techniques. This strengthens the hypothesis that the improvement in problem-solving performance is due to diversity preservation rather than selective pressure.

Our study was limited to a relatively small set of problems, a single GP system (PushGP), and just two down-sampling techniques. 
Future work is needed to verify our findings beyond this context. 
Indeed, many down-sampling techniques have been developed for use in evolutionary computing and machine learning.
Just as there has been recent progress in large-scale benchmarking for selection algorithms \citep{cava2021contemporary,orzechowski_where_2018}, we argue that large-scale benchmarking efforts should be implemented for different down-sampling methods.
Such efforts would help us to disentangle the circumstances where particular down-sampling methods are most appropriate. 
Furthermore, we only studied varied selection pressure and diversity maintenance when used in conjunction with tournament selection. Future work should explore this in conjunction with fitness proportionate selection in order to verify the generality of our conclusions.
Additionally, a more unified theory on the effects of down-sampling on test-based problems could help to tie together disparate results from different application domains in evolutionary computing. 
Finally, future investigations should explore dynamic down-sampling; that is, can we use population statistics to automatically choose and parameterize down-sampling during an evolutionary search?

\section{Acknowledgements}
We thank Charles Ofria, Franz Rothlauf, and the members of the PUSH Lab at Amherst College and UMass Amherst for their helpful discussions and comments.

This work was performed in part using high performance computing equipment obtained under National Science Foundation Grant No. 2117377. Any opinions, findings, and conclusions or recommendations expressed in this publication are those of the authors and do not necessarily reflect the views of the National Science Foundation. This work was performed in part using high performance computing equipment obtained under a grant from the Collaborative R\&D Fund managed by the Massachusetts Technology Collaborative.
%
\bibliographystyle{ACM-Reference-Format}
\bibliography{bib}


\begin{thebibliography}{65}


\ifx \showCODEN    \undefined \def \showCODEN     #1{\unskip}     \fi
\ifx \showDOI      \undefined \def \showDOI       #1{#1}\fi
\ifx \showISBNx    \undefined \def \showISBNx     #1{\unskip}     \fi
\ifx \showISBNxiii \undefined \def \showISBNxiii  #1{\unskip}     \fi
\ifx \showISSN     \undefined \def \showISSN      #1{\unskip}     \fi
\ifx \showLCCN     \undefined \def \showLCCN      #1{\unskip}     \fi
\ifx \shownote     \undefined \def \shownote      #1{#1}          \fi
\ifx \showarticletitle \undefined \def \showarticletitle #1{#1}   \fi
\ifx \showURL      \undefined \def \showURL       {\relax}        \fi
\providecommand\bibfield[2]{#2}
\providecommand\bibinfo[2]{#2}
\providecommand\natexlab[1]{#1}
\providecommand\showeprint[2][]{arXiv:#2}

\bibitem[Aenugu and Spector(2019)]%
        {Aenugu2019}
\bibfield{author}{\bibinfo{person}{Sneha Aenugu} {and} \bibinfo{person}{Lee Spector}.} \bibinfo{year}{2019}\natexlab{}.
\newblock \showarticletitle{Lexicase Selection in Learning Classifier Systems}. In \bibinfo{booktitle}{\emph{Proceedings of the Genetic and Evolutionary Computation Conference}} (Prague, Czech Republic) \emph{(\bibinfo{series}{GECCO '19})}. \bibinfo{publisher}{Association for Computing Machinery}, \bibinfo{address}{New York, NY, USA}, \bibinfo{pages}{356–364}.
\newblock
\showISBNx{9781450361118}
\urldef\tempurl%
\url{https://doi.org/10.1145/3321707.3321828}
\showDOI{\tempurl}


\bibitem[Arabas and Opara(2020)]%
        {Arabas2020PopulationDO}
\bibfield{author}{\bibinfo{person}{Jarosław Arabas} {and} \bibinfo{person}{Karol~R. Opara}.} \bibinfo{year}{2020}\natexlab{}.
\newblock \showarticletitle{Population Diversity of Nonelitist Evolutionary Algorithms in the Exploration Phase}.
\newblock \bibinfo{journal}{\emph{IEEE Transactions on Evolutionary Computation}}  \bibinfo{volume}{24} (\bibinfo{year}{2020}), \bibinfo{pages}{1050--1062}.
\newblock


\bibitem[Blickle and Thiele(1996)]%
        {blickle1996comparison}
\bibfield{author}{\bibinfo{person}{Tobias Blickle} {and} \bibinfo{person}{Lothar Thiele}.} \bibinfo{year}{1996}\natexlab{}.
\newblock \showarticletitle{A comparison of selection schemes used in evolutionary algorithms}.
\newblock \bibinfo{journal}{\emph{Evolutionary Computation}} \bibinfo{volume}{4}, \bibinfo{number}{4} (\bibinfo{year}{1996}), \bibinfo{pages}{361--394}.
\newblock


\bibitem[Boldi et~al\mbox{.}(2023a)]%
        {boldi2023problem}
\bibfield{author}{\bibinfo{person}{Ryan Boldi}, \bibinfo{person}{Ashley Bao}, \bibinfo{person}{Martin Briesch}, \bibinfo{person}{Thomas Helmuth}, \bibinfo{person}{Dominik Sobania}, \bibinfo{person}{Lee Spector}, {and} \bibinfo{person}{Alexander Lalejini}.} \bibinfo{year}{2023}\natexlab{a}.
\newblock \showarticletitle{The Problem Solving Benefits of Down-sampling Vary by Selection Scheme}. In \bibinfo{booktitle}{\emph{Proceedings of the Companion Conference on Genetic and Evolutionary Computation}}. \bibinfo{pages}{527--530}.
\newblock


\bibitem[Boldi et~al\mbox{.}(2024)]%
        {boldi_informed_2023}
\bibfield{author}{\bibinfo{person}{Ryan Boldi}, \bibinfo{person}{Martin Briesch}, \bibinfo{person}{Dominik Sobania}, \bibinfo{person}{Alexander Lalejini}, \bibinfo{person}{Thomas Helmuth}, \bibinfo{person}{Franz Rothlauf}, \bibinfo{person}{Charles Ofria}, {and} \bibinfo{person}{Lee Spector}.} \bibinfo{year}{2024}\natexlab{}.
\newblock \showarticletitle{Informed Down-Sampled Lexicase Selection: Identifying productive training cases for efficient problem solving}.
\newblock \bibinfo{journal}{\emph{Evolutionary Computation}} (\bibinfo{year}{2024}), \bibinfo{pages}{1--32}.
\newblock


\bibitem[Boldi et~al\mbox{.}(2022)]%
        {boldi_2022_environmental}
\bibfield{author}{\bibinfo{person}{Ryan Boldi}, \bibinfo{person}{Thomas Helmuth}, {and} \bibinfo{person}{Lee Spector}.} \bibinfo{year}{2022}\natexlab{}.
\newblock \bibinfo{title}{The Environmental Discontinuity Hypothesis for Down-Sampled Lexicase Selection}.
\newblock
\newblock
\urldef\tempurl%
\url{https://doi.org/10.48550/arxiv.2205.15931}
\showDOI{\tempurl}


\bibitem[Boldi et~al\mbox{.}(2023b)]%
        {boldi_2023_static}
\bibfield{author}{\bibinfo{person}{Ryan Boldi}, \bibinfo{person}{Alexander Lalejini}, \bibinfo{person}{Thomas Helmuth}, {and} \bibinfo{person}{Lee Spector}.} \bibinfo{year}{2023}\natexlab{b}.
\newblock \showarticletitle{A Static Analysis of Informed Down-Samples}. In \bibinfo{booktitle}{\emph{Genetic and Evolutionary Computation Conference Companion (GECCO `23 Companion), July 15--19, 2023, Lisbon, Portugal}}.
\newblock


\bibitem[Briesch et~al\mbox{.}(2023)]%
        {Briesch2023OnTT}
\bibfield{author}{\bibinfo{person}{Martin Briesch}, \bibinfo{person}{Dominik Sobania}, {and} \bibinfo{person}{Franz Rothlauf}.} \bibinfo{year}{2023}\natexlab{}.
\newblock \showarticletitle{On the Trade-Off between Population Size and Number of Generations in GP for Program Synthesis}.
\newblock \bibinfo{journal}{\emph{Proceedings of the Companion Conference on Genetic and Evolutionary Computation}} (\bibinfo{year}{2023}).
\newblock
\urldef\tempurl%
\url{https://api.semanticscholar.org/CorpusID:260119504}
\showURL{%
\tempurl}


\bibitem[Brindle(1980)]%
        {Brindle1980}
\bibfield{author}{\bibinfo{person}{Anne Brindle}.} \bibinfo{year}{1980}\natexlab{}.
\newblock \showarticletitle{Genetic algorithms for function optimization}.
\newblock  (\bibinfo{year}{1980}).
\newblock
\urldef\tempurl%
\url{https://doi.org/10.7939/R3FB4WS2W}
\showDOI{\tempurl}


\bibitem[Butz et~al\mbox{.}(2003)]%
        {Butz2003TournamentSS}
\bibfield{author}{\bibinfo{person}{Martin~Volker Butz}, \bibinfo{person}{Kumara Sastry}, {and} \bibinfo{person}{David~E. Goldberg}.} \bibinfo{year}{2003}\natexlab{}.
\newblock \showarticletitle{Tournament Selection: Stable Fitness Pressure in XCS}. In \bibinfo{booktitle}{\emph{Annual Conference on Genetic and Evolutionary Computation}}.
\newblock


\bibitem[Cava et~al\mbox{.}(2021)]%
        {cava2021contemporary}
\bibfield{author}{\bibinfo{person}{William~La Cava}, \bibinfo{person}{Patryk Orzechowski}, \bibinfo{person}{Bogdan Burlacu}, \bibinfo{person}{Fabricio~Olivetti de Franca}, \bibinfo{person}{Marco Virgolin}, \bibinfo{person}{Ying Jin}, \bibinfo{person}{Michael Kommenda}, {and} \bibinfo{person}{Jason~H. Moore}.} \bibinfo{year}{2021}\natexlab{}.
\newblock \showarticletitle{Contemporary Symbolic Regression Methods and their Relative Performance}. In \bibinfo{booktitle}{\emph{Thirty-fifth Conference on Neural Information Processing Systems Datasets and Benchmarks Track (Round 1)}}.
\newblock
\urldef\tempurl%
\url{https://openreview.net/forum?id=xVQMrDLyGst}
\showURL{%
\tempurl}


\bibitem[Dang et~al\mbox{.}(2019)]%
        {Dang2019RuntimeAO}
\bibfield{author}{\bibinfo{person}{Duc-Cuong Dang}, \bibinfo{person}{Anton~V. Eremeev}, {and} \bibinfo{person}{P. Lehre}.} \bibinfo{year}{2019}\natexlab{}.
\newblock \showarticletitle{Runtime Analysis of Fitness-Proportionate Selection on Linear Functions}.
\newblock \bibinfo{journal}{\emph{ArXiv}}  \bibinfo{volume}{abs/1908.08686} (\bibinfo{year}{2019}).
\newblock


\bibitem[Ding et~al\mbox{.}(2022)]%
        {ding2022scale}
\bibfield{author}{\bibinfo{person}{Li Ding}, \bibinfo{person}{Ryan Boldi}, \bibinfo{person}{Thomas Helmuth}, {and} \bibinfo{person}{Lee Spector}.} \bibinfo{year}{2022}\natexlab{}.
\newblock \showarticletitle{Lexicase Selection at Scale}. In \bibinfo{booktitle}{\emph{Genetic and Evolutionary Computation Conference Companion (GECCO '22 Companion), July 9--13, 2022, Boston, MA, USA}}.
\newblock


\bibitem[Ding and Spector(2021)]%
        {Ding2022optimizing}
\bibfield{author}{\bibinfo{person}{Li Ding} {and} \bibinfo{person}{Lee Spector}.} \bibinfo{year}{2021}\natexlab{}.
\newblock \showarticletitle{Optimizing neural networks with gradient lexicase selection}. In \bibinfo{booktitle}{\emph{International Conference on Learning Representations}}.
\newblock


\bibitem[Dolson and Ofria(2021)]%
        {dolson_digital_2021}
\bibfield{author}{\bibinfo{person}{Emily Dolson} {and} \bibinfo{person}{Charles Ofria}.} \bibinfo{year}{2021}\natexlab{}.
\newblock \showarticletitle{Digital {Evolution} for {Ecology} {Research}: {A} {Review}}.
\newblock \bibinfo{journal}{\emph{Frontiers in Ecology and Evolution}}  \bibinfo{volume}{9} (\bibinfo{year}{2021}), \bibinfo{pages}{18}.
\newblock
\urldef\tempurl%
\url{https://doi.org/10.3389/fevo.2021.750779}
\showDOI{\tempurl}


\bibitem[Dolson et~al\mbox{.}(2018)]%
        {dolson_ecological_2018}
\bibfield{author}{\bibinfo{person}{Emily~L Dolson}, \bibinfo{person}{Wolfgang Banzhaf}, {and} \bibinfo{person}{Charles Ofria}.} \bibinfo{year}{2018}\natexlab{}.
\newblock \bibinfo{booktitle}{\emph{Ecological theory provides insights about evolutionary computation}}.
\newblock \bibinfo{type}{preprint}. \bibinfo{institution}{PeerJ Preprints}.
\newblock
\urldef\tempurl%
\url{https://doi.org/10.7287/peerj.preprints.27315v1}
\showDOI{\tempurl}


\bibitem[Fang and Li(2010)]%
        {fang_review_2010}
\bibfield{author}{\bibinfo{person}{Yongsheng Fang} {and} \bibinfo{person}{Jun Li}.} \bibinfo{year}{2010}\natexlab{}.
\newblock \showarticletitle{A {Review} of {Tournament} {Selection} in {Genetic} {Programming}}. In \bibinfo{booktitle}{\emph{Advances in {Computation} and {Intelligence}}} \emph{(\bibinfo{series}{Lecture {Notes} in {Computer} {Science}})}, \bibfield{editor}{\bibinfo{person}{Zhihua Cai}, \bibinfo{person}{Chengyu Hu}, \bibinfo{person}{Zhuo Kang}, {and} \bibinfo{person}{Yong Liu}} (Eds.). \bibinfo{publisher}{Springer}, \bibinfo{address}{Berlin, Heidelberg}, \bibinfo{pages}{181--192}.
\newblock
\showISBNx{978-3-642-16493-4}
\urldef\tempurl%
\url{https://doi.org/10.1007/978-3-642-16493-4-19}
\showDOI{\tempurl}


\bibitem[Ferguson et~al\mbox{.}(2019)]%
        {Ofria:2019:GPTP}
\bibfield{author}{\bibinfo{person}{Austin~J. Ferguson}, \bibinfo{person}{Jose~Guadalupe Hernandez}, \bibinfo{person}{Daniel Junghans}, \bibinfo{person}{Alexander Lalejini}, \bibinfo{person}{Emily Dolson}, {and} \bibinfo{person}{Charles Ofria}.} \bibinfo{year}{2019}\natexlab{}.
\newblock \showarticletitle{Characterizing the effects of random subsampling and dilution on Lexicase selection}. In \bibinfo{booktitle}{\emph{Genetic Programming Theory and Practice XVII}}, \bibfield{editor}{\bibinfo{person}{Wolfgang Banzhaf}, \bibinfo{person}{Erik Goodman}, \bibinfo{person}{Leigh Sheneman}, \bibinfo{person}{Leonardo Trujillo}, {and} \bibinfo{person}{Bill Worzel}} (Eds.). \bibinfo{publisher}{Springer}, \bibinfo{address}{East Lansing, MI, USA}, \bibinfo{pages}{1--23}.
\newblock
\urldef\tempurl%
\url{https://doi.org/doi:10.1007/978-3-030-39958-0-1}
\showDOI{\tempurl}


\bibitem[Gathercole and Ross(1994)]%
        {ga94aGathercole}
\bibfield{author}{\bibinfo{person}{Chris Gathercole} {and} \bibinfo{person}{Peter Ross}.} \bibinfo{year}{1994}\natexlab{}.
\newblock \showarticletitle{Dynamic Training Subset Selection for Supervised Learning in Genetic Programming}. In \bibinfo{booktitle}{\emph{Parallel Problem Solving from Nature III}} \emph{(\bibinfo{series}{LNCS}, Vol.~\bibinfo{volume}{866})}, \bibfield{editor}{\bibinfo{person}{Yuval Davidor}, \bibinfo{person}{Hans-Paul Schwefel}, {and} \bibinfo{person}{Reinhard M{\"a}nner}} (Eds.). \bibinfo{publisher}{Springer-Verlag}, \bibinfo{address}{Jerusalem}, \bibinfo{pages}{312--321}.
\newblock
\showISBNx{3-540-58484-6}
\urldef\tempurl%
\url{https://doi.org/doi:10.1007/3-540-58484-6-275}
\showDOI{\tempurl}


\bibitem[Geiger et~al\mbox{.}(2023)]%
        {geiger2023}
\bibfield{author}{\bibinfo{person}{Alina Geiger}, \bibinfo{person}{Dominik Sobania}, {and} \bibinfo{person}{Franz Rothlauf}.} \bibinfo{year}{2023}\natexlab{}.
\newblock \showarticletitle{Down-Sampled Epsilon-Lexicase Selection for Real-World Symbolic Regression Problems}. In \bibinfo{booktitle}{\emph{Proceedings of the Genetic and Evolutionary Computation Conference}} (Lisbon, Portugal) \emph{(\bibinfo{series}{GECCO '23})}. \bibinfo{publisher}{Association for Computing Machinery}, \bibinfo{address}{New York, NY, USA}, \bibinfo{pages}{1109–1117}.
\newblock
\showISBNx{9798400701191}
\urldef\tempurl%
\url{https://doi.org/10.1145/3583131.3590400}
\showDOI{\tempurl}


\bibitem[Goldberg and Deb(1991)]%
        {goldberg_comparative_1991}
\bibfield{author}{\bibinfo{person}{David~E. Goldberg} {and} \bibinfo{person}{Kalyanmoy Deb}.} \bibinfo{year}{1991}\natexlab{}.
\newblock \showarticletitle{A {Comparative} {Analysis} of {Selection} {Schemes} {Used} in {Genetic} {Algorithms}}.
\newblock In \bibinfo{booktitle}{\emph{Foundations of {Genetic} {Algorithms}}}. Vol.~\bibinfo{volume}{1}. \bibinfo{publisher}{Elsevier}, \bibinfo{pages}{69--93}.
\newblock
\showISBNx{978-0-08-050684-5}
\urldef\tempurl%
\url{https://doi.org/10.1016/B978-0-08-050684-5.50008-2}
\showDOI{\tempurl}


\bibitem[Goncalves and Silva(2013)]%
        {goncalves:2013:EuroGP}
\bibfield{author}{\bibinfo{person}{Ivo Goncalves} {and} \bibinfo{person}{Sara Silva}.} \bibinfo{year}{2013}\natexlab{}.
\newblock \showarticletitle{Balancing Learning and Overfitting in Genetic Programming with Interleaved Sampling of Training data}. In \bibinfo{booktitle}{\emph{Proceedings of the 16th European Conference on Genetic Programming, EuroGP 2013}} \emph{(\bibinfo{series}{LNCS}, Vol.~\bibinfo{volume}{7831})}, \bibfield{editor}{\bibinfo{person}{Krzysztof Krawiec}, \bibinfo{person}{Alberto Moraglio}, \bibinfo{person}{Ting Hu}, \bibinfo{person}{A.~Sima Uyar}, {and} \bibinfo{person}{Bin Hu}} (Eds.). \bibinfo{publisher}{Springer Verlag}, \bibinfo{address}{Vienna, Austria}, \bibinfo{pages}{73--84}.
\newblock
\urldef\tempurl%
\url{https://doi.org/doi:10.1007/978-3-642-37207-0_7}
\showDOI{\tempurl}


\bibitem[Gonçalves et~al\mbox{.}(2012)]%
        {goncalves_random_2012}
\bibfield{author}{\bibinfo{person}{Ivo Gonçalves}, \bibinfo{person}{Sara Silva}, \bibinfo{person}{Joana B.~Melo}, {and} \bibinfo{person}{Joao Carreiras}.} \bibinfo{year}{2012}\natexlab{}.
\newblock \bibinfo{booktitle}{\emph{Random {Sampling} {Technique} for {Overfitting} {Control} in {Genetic} {Programming}}}.
\newblock
\showISBNx{978-3-642-29138-8}
\urldef\tempurl%
\url{https://doi.org/10.1007/978-3-642-29139-5_19}
\showDOI{\tempurl}
\newblock
\shownote{Pages: 229}.


\bibitem[Helmuth and Kelly(2021)]%
        {helmuth_psb2_2021}
\bibfield{author}{\bibinfo{person}{Thomas Helmuth} {and} \bibinfo{person}{Peter Kelly}.} \bibinfo{year}{2021}\natexlab{}.
\newblock \showarticletitle{{PSB2}: the second program synthesis benchmark suite}. In \bibinfo{booktitle}{\emph{Proceedings of the {Genetic} and {Evolutionary} {Computation} {Conference}}}. \bibinfo{publisher}{ACM}, \bibinfo{address}{Lille France}, \bibinfo{pages}{785--794}.
\newblock
\showISBNx{978-1-4503-8350-9}
\urldef\tempurl%
\url{https://doi.org/10.1145/3449639.3459285}
\showDOI{\tempurl}


\bibitem[Helmuth and Kelly(2022)]%
        {helmuth_applying_2022}
\bibfield{author}{\bibinfo{person}{Thomas Helmuth} {and} \bibinfo{person}{Peter Kelly}.} \bibinfo{year}{2022}\natexlab{}.
\newblock \showarticletitle{Applying genetic programming to {PSB2}: the next generation program synthesis benchmark suite}.
\newblock \bibinfo{journal}{\emph{Genetic Programming and Evolvable Machines}} (\bibinfo{date}{June} \bibinfo{year}{2022}).
\newblock
\showISSN{1389-2576, 1573-7632}
\urldef\tempurl%
\url{https://doi.org/10.1007/s10710-022-09434-y}
\showDOI{\tempurl}


\bibitem[Helmuth et~al\mbox{.}(2022)]%
        {helmuth_population_2022}
\bibfield{author}{\bibinfo{person}{Thomas Helmuth}, \bibinfo{person}{Johannes Lengler}, {and} \bibinfo{person}{William La~Cava}.} \bibinfo{year}{2022}\natexlab{}.
\newblock \showarticletitle{Population {Diversity} {Leads} to {Short} {Running} {Times} of {Lexicase} {Selection}}. In \bibinfo{booktitle}{\emph{Parallel {Problem} {Solving} from {Nature} – {PPSN} {XVII}}}, \bibfield{editor}{\bibinfo{person}{Günter Rudolph}, \bibinfo{person}{Anna~V. Kononova}, \bibinfo{person}{Hernán Aguirre}, \bibinfo{person}{Pascal Kerschke}, \bibinfo{person}{Gabriela Ochoa}, {and} \bibinfo{person}{Tea Tušar}} (Eds.). \bibinfo{publisher}{Springer International Publishing}, \bibinfo{address}{Cham}, \bibinfo{pages}{485--498}.
\newblock
\showISBNx{978-3-031-14721-0}


\bibitem[Helmuth et~al\mbox{.}(2016a)]%
        {helmuth_effects_2016}
\bibfield{author}{\bibinfo{person}{Thomas Helmuth}, \bibinfo{person}{Nicholas~Freitag McPhee}, {and} \bibinfo{person}{Lee Spector}.} \bibinfo{year}{2016}\natexlab{a}.
\newblock \showarticletitle{Effects of {Lexicase} and {Tournament} {Selection} on {Diversity} {Recovery} and {Maintenance}}. In \bibinfo{booktitle}{\emph{Proceedings of the 2016 on {Genetic} and {Evolutionary} {Computation} {Conference} {Companion}}} \emph{(\bibinfo{series}{{GECCO} '16 {Companion}})}. \bibinfo{publisher}{Association for Computing Machinery}, \bibinfo{address}{New York, NY, USA}, \bibinfo{pages}{983--990}.
\newblock
\showISBNx{978-1-4503-4323-7}
\urldef\tempurl%
\url{https://doi.org/10.1145/2908961.2931657}
\showDOI{\tempurl}


\bibitem[Helmuth et~al\mbox{.}(2016b)]%
        {riolo_lexicase_2016}
\bibfield{author}{\bibinfo{person}{Thomas Helmuth}, \bibinfo{person}{Nicholas~Freitag McPhee}, {and} \bibinfo{person}{Lee Spector}.} \bibinfo{year}{2016}\natexlab{b}.
\newblock \showarticletitle{Lexicase {Selection} for {Program} {Synthesis}: {A} {Diversity} {Analysis}}.
\newblock In \bibinfo{booktitle}{\emph{Genetic {Programming} {Theory} and {Practice} {XIII}}}, \bibfield{editor}{\bibinfo{person}{Rick Riolo}, \bibinfo{person}{W.P. Worzel}, \bibinfo{person}{Mark Kotanchek}, {and} \bibinfo{person}{Arthur Kordon}} (Eds.). \bibinfo{publisher}{Springer International Publishing}, \bibinfo{address}{Cham}, \bibinfo{pages}{151--167}.
\newblock
\showISBNx{978-3-319-34221-4 978-3-319-34223-8}
\urldef\tempurl%
\url{https://doi.org/10.1007/978-3-319-34223-8\_9}
\showDOI{\tempurl}
\newblock
\shownote{Series Title: Genetic and Evolutionary Computation}.


\bibitem[Helmuth et~al\mbox{.}(2020)]%
        {helmuth_importance_2020}
\bibfield{author}{\bibinfo{person}{Thomas Helmuth}, \bibinfo{person}{Edward Pantridge}, {and} \bibinfo{person}{Lee Spector}.} \bibinfo{year}{2020}\natexlab{}.
\newblock \showarticletitle{On the importance of specialists for lexicase selection}.
\newblock \bibinfo{journal}{\emph{Genetic Programming and Evolvable Machines}} \bibinfo{volume}{21}, \bibinfo{number}{3} (\bibinfo{date}{Sept.} \bibinfo{year}{2020}), \bibinfo{pages}{349--373}.
\newblock
\showISSN{1389-2576, 1573-7632}
\urldef\tempurl%
\url{https://doi.org/10.1007/s10710-020-09377-2}
\showDOI{\tempurl}


\bibitem[Helmuth and Spector(2015)]%
        {helmuth_general_2015}
\bibfield{author}{\bibinfo{person}{Thomas Helmuth} {and} \bibinfo{person}{Lee Spector}.} \bibinfo{year}{2015}\natexlab{}.
\newblock \showarticletitle{General {Program} {Synthesis} {Benchmark} {Suite}}. In \bibinfo{booktitle}{\emph{Proceedings of the 2015 {Annual} {Conference} on {Genetic} and {Evolutionary} {Computation}}}. \bibinfo{publisher}{ACM}, \bibinfo{address}{Madrid Spain}, \bibinfo{pages}{1039--1046}.
\newblock
\showISBNx{978-1-4503-3472-3}
\urldef\tempurl%
\url{https://doi.org/10.1145/2739480.2754769}
\showDOI{\tempurl}


\bibitem[Helmuth and Spector(2021)]%
        {Helmuth2021benefits}
\bibfield{author}{\bibinfo{person}{Thomas Helmuth} {and} \bibinfo{person}{Lee Spector}.} \bibinfo{year}{2021}\natexlab{}.
\newblock \showarticletitle{{Problem-solving benefits of down-sampled lexicase selection}}.
\newblock \bibinfo{journal}{\emph{Artificial Life}} (\bibinfo{date}{jun} \bibinfo{year}{2021}), \bibinfo{pages}{1--21}.
\newblock
\showISSN{1530-9185}
\urldef\tempurl%
\url{https://doi.org/10.1162/artl\_a\_00341}
\showDOI{\tempurl}
\showeprint[arxiv]{2106.06085}


\bibitem[Helmuth et~al\mbox{.}(2015)]%
        {Helmuth_solving_2015}
\bibfield{author}{\bibinfo{person}{Thomas Helmuth}, \bibinfo{person}{Lee Spector}, {and} \bibinfo{person}{James Matheson}.} \bibinfo{year}{2015}\natexlab{}.
\newblock \showarticletitle{Solving Uncompromising Problems With Lexicase Selection}.
\newblock \bibinfo{journal}{\emph{IEEE Transactions on Evolutionary Computation}} \bibinfo{volume}{19}, \bibinfo{number}{5} (\bibinfo{year}{2015}), \bibinfo{pages}{630--643}.
\newblock
\urldef\tempurl%
\url{https://doi.org/10.1109/TEVC.2014.2362729}
\showDOI{\tempurl}


\bibitem[Hernandez et~al\mbox{.}(2022a)]%
        {hernandez_what_2022}
\bibfield{author}{\bibinfo{person}{Jose~Guadalupe Hernandez}, \bibinfo{person}{Alexander Lalejini}, {and} \bibinfo{person}{Emily Dolson}.} \bibinfo{year}{2022}\natexlab{a}.
\newblock \showarticletitle{What {Can} {Phylogenetic} {Metrics} {Tell} us {About} {Useful} {Diversity} in {Evolutionary} {Algorithms}?}
\newblock In \bibinfo{booktitle}{\emph{Genetic {Programming} {Theory} and {Practice} {XVIII}}}, \bibfield{editor}{\bibinfo{person}{Wolfgang Banzhaf}, \bibinfo{person}{Leonardo Trujillo}, \bibinfo{person}{Stephan Winkler}, {and} \bibinfo{person}{Bill Worzel}} (Eds.). \bibinfo{publisher}{Springer Nature Singapore}, \bibinfo{address}{Singapore}, \bibinfo{pages}{63--82}.
\newblock
\showISBNx{9789811681127 9789811681134}
\urldef\tempurl%
\url{https://doi.org/10.1007/978-981-16-8113-4\_4}
\showDOI{\tempurl}
\newblock
\shownote{Series Title: Genetic and Evolutionary Computation}.


\bibitem[Hernandez et~al\mbox{.}(2019)]%
        {hernandez_random_2019}
\bibfield{author}{\bibinfo{person}{Jose~Guadalupe Hernandez}, \bibinfo{person}{Alexander Lalejini}, \bibinfo{person}{Emily Dolson}, {and} \bibinfo{person}{Charles Ofria}.} \bibinfo{year}{2019}\natexlab{}.
\newblock \showarticletitle{Random subsampling improves performance in lexicase selection}. In \bibinfo{booktitle}{\emph{Proceedings of the {Genetic} and {Evolutionary} {Computation} {Conference} {Companion}}}. \bibinfo{publisher}{ACM}, \bibinfo{address}{Prague Czech Republic}, \bibinfo{pages}{2028--2031}.
\newblock
\showISBNx{978-1-4503-6748-6}
\urldef\tempurl%
\url{https://doi.org/10.1145/3319619.3326900}
\showDOI{\tempurl}


\bibitem[Hernandez et~al\mbox{.}(2022b)]%
        {Hernandez2022}
\bibfield{author}{\bibinfo{person}{Jose~Guadalupe Hernandez}, \bibinfo{person}{Alexander Lalejini}, {and} \bibinfo{person}{Charles Ofria}.} \bibinfo{year}{2022}\natexlab{b}.
\newblock \bibinfo{booktitle}{\emph{An Exploration of Exploration: Measuring the Ability of Lexicase Selection to Find Obscure Pathways to Optimality}}.
\newblock \bibinfo{publisher}{Springer Nature Singapore}, \bibinfo{address}{Singapore}, \bibinfo{pages}{83--107}.
\newblock
\showISBNx{978-981-16-8113-4}
\urldef\tempurl%
\url{https://doi.org/10.1007/978-981-16-8113-4\_5}
\showDOI{\tempurl}


\bibitem[Hernandez et~al\mbox{.}(2022c)]%
        {hernandez_diagnostics_2022}
\bibfield{author}{\bibinfo{person}{Jose~Guadalupe Hernandez}, \bibinfo{person}{Alexander Lalejini}, {and} \bibinfo{person}{Charles Ofria}.} \bibinfo{year}{2022}\natexlab{c}.
\newblock \bibinfo{title}{A suite of diagnostic metrics for characterizing selection schemes}.
\newblock
\newblock
\urldef\tempurl%
\url{https://doi.org/10.48550/ARXIV.2204.13839}
\showDOI{\tempurl}


\bibitem[Hochbaum and Shmoys(1985)]%
        {Hochbaum1985ABP}
\bibfield{author}{\bibinfo{person}{Dorit~S. Hochbaum} {and} \bibinfo{person}{David~B. Shmoys}.} \bibinfo{year}{1985}\natexlab{}.
\newblock \showarticletitle{A Best Possible Heuristic for the k-Center Problem}.
\newblock \bibinfo{journal}{\emph{Math. Oper. Res.}}  \bibinfo{volume}{10} (\bibinfo{year}{1985}), \bibinfo{pages}{180--184}.
\newblock


\bibitem[Holland(1992)]%
        {holland_adaptation_1992}
\bibfield{author}{\bibinfo{person}{John~H. Holland}.} \bibinfo{year}{1992}\natexlab{}.
\newblock \bibinfo{booktitle}{\emph{Adaptation in natural and artificial systems: an introductory analysis with applications to biology, control, and artificial intelligence} (\bibinfo{edition}{1st mit press ed} ed.)}.
\newblock \bibinfo{publisher}{MIT Press}, \bibinfo{address}{Cambridge, Mass}.
\newblock
\showISBNx{978-0-262-08213-6 978-0-262-58111-0}


\bibitem[Hornby(2006)]%
        {hornby_alps_2006}
\bibfield{author}{\bibinfo{person}{Gregory~S. Hornby}.} \bibinfo{year}{2006}\natexlab{}.
\newblock \showarticletitle{{ALPS}: the age-layered population structure for reducing the problem of premature convergence}. In \bibinfo{booktitle}{\emph{Proceedings of the 8th annual conference on {Genetic} and evolutionary computation - {GECCO} '06}}. \bibinfo{publisher}{ACM Press}, \bibinfo{address}{Seattle, Washington, USA}, \bibinfo{pages}{815}.
\newblock
\showISBNx{978-1-59593-186-3}
\urldef\tempurl%
\url{https://doi.org/10.1145/1143997.1144142}
\showDOI{\tempurl}


\bibitem[Krawiec and Nawrocki(2013)]%
        {Krawiec13IFS}
\bibfield{author}{\bibinfo{person}{Krzysztof Krawiec} {and} \bibinfo{person}{Mateusz Nawrocki}.} \bibinfo{year}{2013}\natexlab{}.
\newblock \showarticletitle{Implicit Fitness Sharing for Evolutionary Synthesis of License Plate Detectors}. In \bibinfo{booktitle}{\emph{Applications of Evolutionary Computation}}, \bibfield{editor}{\bibinfo{person}{Anna~I. Esparcia-Alc{\'a}zar}} (Ed.). \bibinfo{publisher}{Springer Berlin Heidelberg}, \bibinfo{address}{Berlin, Heidelberg}, \bibinfo{pages}{376--386}.
\newblock
\showISBNx{978-3-642-37192-9}


\bibitem[Lalejini et~al\mbox{.}(2022)]%
        {lalejini_artificial_2022}
\bibfield{author}{\bibinfo{person}{Alexander Lalejini}, \bibinfo{person}{Emily Dolson}, \bibinfo{person}{Anya~E Vostinar}, {and} \bibinfo{person}{Luis Zaman}.} \bibinfo{year}{2022}\natexlab{}.
\newblock \showarticletitle{Artificial selection methods from evolutionary computing show promise for directed evolution of microbes}.
\newblock \bibinfo{journal}{\emph{eLife}}  \bibinfo{volume}{11} (\bibinfo{date}{Aug.} \bibinfo{year}{2022}), \bibinfo{pages}{e79665}.
\newblock
\showISSN{2050-084X}
\urldef\tempurl%
\url{https://doi.org/10.7554/eLife.79665}
\showDOI{\tempurl}


\bibitem[Lalejini et~al\mbox{.}(2021)]%
        {lalejini_adaptive_2021}
\bibfield{author}{\bibinfo{person}{Alexander Lalejini}, \bibinfo{person}{Austin~J. Ferguson}, \bibinfo{person}{Nkrumah~A. Grant}, {and} \bibinfo{person}{Charles Ofria}.} \bibinfo{year}{2021}\natexlab{}.
\newblock \showarticletitle{Adaptive {Phenotypic} {Plasticity} {Stabilizes} {Evolution} in {Fluctuating} {Environments}}.
\newblock \bibinfo{journal}{\emph{Frontiers in Ecology and Evolution}}  \bibinfo{volume}{9} (\bibinfo{date}{Aug.} \bibinfo{year}{2021}), \bibinfo{pages}{715381}.
\newblock
\showISSN{2296-701X}
\urldef\tempurl%
\url{https://doi.org/10.3389/fevo.2021.715381}
\showDOI{\tempurl}


\bibitem[Lalejini et~al\mbox{.}(2023)]%
        {Lalejini2023PhylogenyinformedFE}
\bibfield{author}{\bibinfo{person}{Alexander Lalejini}, \bibinfo{person}{Matthew~Andres Moreno}, \bibinfo{person}{Jose~Guadalupe Hernandez}, {and} \bibinfo{person}{Emily Dolson}.} \bibinfo{year}{2023}\natexlab{}.
\newblock \showarticletitle{Phylogeny-informed fitness estimation}.
\newblock \bibinfo{journal}{\emph{ArXiv}}  \bibinfo{volume}{abs/2306.03970} (\bibinfo{year}{2023}).
\newblock
\urldef\tempurl%
\url{https://api.semanticscholar.org/CorpusID:259095608}
\showURL{%
\tempurl}


\bibitem[Langdon(2011)]%
        {langdon_minimising_2011}
\bibfield{author}{\bibinfo{person}{W. Langdon}.} \bibinfo{year}{2011}\natexlab{}.
\newblock \showarticletitle{Minimising testing in genetic programming}.
\newblock \bibinfo{journal}{\emph{RN}} \bibinfo{volume}{11}, \bibinfo{number}{10} (\bibinfo{year}{2011}).
\newblock


\bibitem[Lasarczyk et~al\mbox{.}(2004)]%
        {lasarczyk_dynamic_2004}
\bibfield{author}{\bibinfo{person}{Christian~W.G. Lasarczyk}, \bibinfo{person}{Peter Dittrich}, {and} \bibinfo{person}{Wolfgang Banzhaf}.} \bibinfo{year}{2004}\natexlab{}.
\newblock \showarticletitle{Dynamic {Subset} {Selection} {Based} on a {Fitness} {Case} {Topology}}.
\newblock \bibinfo{journal}{\emph{Evolutionary Computation}} \bibinfo{volume}{12}, \bibinfo{number}{2} (\bibinfo{date}{June} \bibinfo{year}{2004}), \bibinfo{pages}{223--242}.
\newblock
\showISSN{1063-6560, 1530-9304}
\urldef\tempurl%
\url{https://doi.org/10.1162/106365604773955157}
\showDOI{\tempurl}


\bibitem[Lenski et~al\mbox{.}(2003)]%
        {lenski_evolutionary_2003}
\bibfield{author}{\bibinfo{person}{Richard~E. Lenski}, \bibinfo{person}{Charles Ofria}, \bibinfo{person}{Robert~T. Pennock}, {and} \bibinfo{person}{Christoph Adami}.} \bibinfo{year}{2003}\natexlab{}.
\newblock \showarticletitle{The evolutionary origin of complex features}.
\newblock \bibinfo{journal}{\emph{Nature}} \bibinfo{volume}{423}, \bibinfo{number}{6936} (\bibinfo{date}{May} \bibinfo{year}{2003}), \bibinfo{pages}{139--144}.
\newblock
\showISSN{0028-0836, 1476-4687}
\urldef\tempurl%
\url{https://doi.org/10.1038/nature01568}
\showDOI{\tempurl}


\bibitem[Liu and Khoshgoftaar(2004)]%
        {liu_reducing_2004}
\bibfield{author}{\bibinfo{person}{Yi Liu} {and} \bibinfo{person}{Taghi Khoshgoftaar}.} \bibinfo{year}{2004}\natexlab{}.
\newblock \showarticletitle{Reducing overfitting in genetic programming models for software quality classification}. In \bibinfo{booktitle}{\emph{Proceedings of the {Eighth} {IEEE} international conference on {High} assurance systems engineering}} \emph{(\bibinfo{series}{{HASE}'04})}. \bibinfo{publisher}{IEEE Computer Society}, \bibinfo{address}{USA}, \bibinfo{pages}{56--65}.
\newblock
\showISBNx{978-0-7695-2094-0}


\bibitem[McKay(2000)]%
        {ifsGPMcKay00}
\bibfield{author}{\bibinfo{person}{R~I~(Bob) McKay}.} \bibinfo{year}{2000}\natexlab{}.
\newblock \showarticletitle{Fitness Sharing in Genetic Programming}. In \bibinfo{booktitle}{\emph{Proceedings of the 2nd Annual Conference on Genetic and Evolutionary Computation}} (Las Vegas, Nevada) \emph{(\bibinfo{series}{GECCO'00})}. \bibinfo{publisher}{Morgan Kaufmann Publishers Inc.}, \bibinfo{address}{San Francisco, CA, USA}, \bibinfo{pages}{435–442}.
\newblock
\showISBNx{1558607080}


\bibitem[Metevier et~al\mbox{.}(2019)]%
        {metevier_lexicase_2019}
\bibfield{author}{\bibinfo{person}{Blossom Metevier}, \bibinfo{person}{Anil~Kumar Saini}, {and} \bibinfo{person}{Lee Spector}.} \bibinfo{year}{2019}\natexlab{}.
\newblock \showarticletitle{Lexicase {Selection} {Beyond} {Genetic} {Programming}}.
\newblock In \bibinfo{booktitle}{\emph{Genetic {Programming} {Theory} and {Practice} {XVI}}}, \bibfield{editor}{\bibinfo{person}{Wolfgang Banzhaf}, \bibinfo{person}{Lee Spector}, {and} \bibinfo{person}{Leigh Sheneman}} (Eds.). \bibinfo{publisher}{Springer International Publishing}, \bibinfo{address}{Cham}, \bibinfo{pages}{123--136}.
\newblock
\showISBNx{978-3-030-04735-1}
\urldef\tempurl%
\url{https://doi.org/10.1007/978-3-030-04735-1\_7}
\showDOI{\tempurl}


\bibitem[Moore and Stanton(2017)]%
        {moore2017lex}
\bibfield{author}{\bibinfo{person}{Jared~M. Moore} {and} \bibinfo{person}{Adam Stanton}.} \bibinfo{year}{2017}\natexlab{}.
\newblock \showarticletitle{Lexicase selection outperforms previous strategies for incremental evolution of virtual creature controllers}. In \bibinfo{booktitle}{\emph{Proceedings of the Fourteenth European Conference Artificial Life, ECAL 2017, Lyon, France, September 4-8, 2017}}, \bibfield{editor}{\bibinfo{person}{Carole Knibbe}, \bibinfo{person}{Guillaume Beslon}, \bibinfo{person}{David~P. Parsons}, \bibinfo{person}{Dusan Misevic}, \bibinfo{person}{Jonathan Rouzaud-Cornabas}, \bibinfo{person}{Nicolas Bred{\`e}che}, \bibinfo{person}{Salima Hassas}, \bibinfo{person}{Olivier~Simonin 0001}, {and} \bibinfo{person}{H{\'e}di Soula}} (Eds.). \bibinfo{publisher}{MIT Press}, \bibinfo{pages}{290--297}.
\newblock
\showISBNx{978-0-262-34633-7}
\urldef\tempurl%
\url{http://cognet.mit.edu/journal/ecal2017}
\showURL{%
\tempurl}


\bibitem[Mouret and Clune(2015)]%
        {Mouret2015IlluminatingSS}
\bibfield{author}{\bibinfo{person}{Jean-Baptiste Mouret} {and} \bibinfo{person}{Jeff Clune}.} \bibinfo{year}{2015}\natexlab{}.
\newblock \showarticletitle{Illuminating search spaces by mapping elites}.
\newblock \bibinfo{journal}{\emph{ArXiv}}  \bibinfo{volume}{abs/1504.04909} (\bibinfo{year}{2015}).
\newblock
\urldef\tempurl%
\url{https://api.semanticscholar.org/CorpusID:14759751}
\showURL{%
\tempurl}


\bibitem[Nordin and Banzhaf(1997)]%
        {nordin_-line_1997}
\bibfield{author}{\bibinfo{person}{Peter Nordin} {and} \bibinfo{person}{Wolfgang Banzhaf}.} \bibinfo{year}{1997}\natexlab{}.
\newblock \showarticletitle{An {On}-{Line} {Method} to {Evolve} {Behavior} and to {Control} a {Miniature} {Robot} in {Real} {Time} with {Genetic} {Programming}}.
\newblock \bibinfo{journal}{\emph{Adaptive Behavior}} \bibinfo{volume}{5}, \bibinfo{number}{2} (\bibinfo{date}{Jan.} \bibinfo{year}{1997}), \bibinfo{pages}{107--140}.
\newblock
\showISSN{1059-7123, 1741-2633}
\urldef\tempurl%
\url{https://doi.org/10.1177/105971239700500201}
\showDOI{\tempurl}


\bibitem[Orzechowski et~al\mbox{.}(2018)]%
        {orzechowski_where_2018}
\bibfield{author}{\bibinfo{person}{Patryk Orzechowski}, \bibinfo{person}{William La~Cava}, {and} \bibinfo{person}{Jason~H. Moore}.} \bibinfo{year}{2018}\natexlab{}.
\newblock \showarticletitle{Where are we now? a large benchmark study of recent symbolic regression methods}. In \bibinfo{booktitle}{\emph{Proceedings of the {Genetic} and {Evolutionary} {Computation} {Conference}}}. \bibinfo{publisher}{ACM}, \bibinfo{address}{Kyoto Japan}, \bibinfo{pages}{1183--1190}.
\newblock
\showISBNx{978-1-4503-5618-3}
\urldef\tempurl%
\url{https://doi.org/10.1145/3205455.3205539}
\showDOI{\tempurl}


\bibitem[Ross(1999)]%
        {ross_effects_1999}
\bibfield{author}{\bibinfo{person}{Brian~J Ross}.} \bibinfo{year}{1999}\natexlab{}.
\newblock \bibinfo{booktitle}{\emph{The {Effects} of {Randomly} {Sampled} {Training} {Data} on {Program} {Evolution}}}.
\newblock \bibinfo{type}{Technical {Report}} CS-99-03. \bibinfo{institution}{Dept. of Computer Science, Brock University}, \bibinfo{address}{Canada}.
\newblock


\bibitem[Schweim et~al\mbox{.}(2022)]%
        {schweim_effects_2022}
\bibfield{author}{\bibinfo{person}{Dirk Schweim}, \bibinfo{person}{Dominik Sobania}, {and} \bibinfo{person}{Franz Rothlauf}.} \bibinfo{year}{2022}\natexlab{}.
\newblock \showarticletitle{Effects of the Training Set Size: A Comparison of Standard and Down-Sampled Lexicase Selection in Program Synthesis}. In \bibinfo{booktitle}{\emph{2022 IEEE Congress on Evolutionary Computation (CEC)}}. \bibinfo{pages}{1--8}.
\newblock
\urldef\tempurl%
\url{https://doi.org/10.1109/CEC55065.2022.9870337}
\showDOI{\tempurl}


\bibitem[Shahbandegan et~al\mbox{.}(2022)]%
        {shahbandegan_untangling_2022}
\bibfield{author}{\bibinfo{person}{Shakiba Shahbandegan}, \bibinfo{person}{Jose~Guadalupe Hernandez}, \bibinfo{person}{Alexander Lalejini}, {and} \bibinfo{person}{Emily Dolson}.} \bibinfo{year}{2022}\natexlab{}.
\newblock \showarticletitle{Untangling phylogenetic diversity's role in evolutionary computation using a suite of diagnostic fitness landscapes}. In \bibinfo{booktitle}{\emph{Proceedings of the {Genetic} and {Evolutionary} {Computation} {Conference} {Companion}}}. \bibinfo{publisher}{ACM}, \bibinfo{address}{Boston Massachusetts}, \bibinfo{pages}{2322--2325}.
\newblock
\showISBNx{978-1-4503-9268-6}
\urldef\tempurl%
\url{https://doi.org/10.1145/3520304.3534028}
\showDOI{\tempurl}


\bibitem[Smith et~al\mbox{.}(1993)]%
        {whitley93IFS}
\bibfield{author}{\bibinfo{person}{Robert~E. Smith}, \bibinfo{person}{Stephanie Forrest}, {and} \bibinfo{person}{Alan~S. Perelson}.} \bibinfo{year}{1993}\natexlab{}.
\newblock \showarticletitle{Population Diversity in an Immune System Model: Implications for Genetic Search}.
\newblock In \bibinfo{booktitle}{\emph{Foundations of Genetic Algorithms}}, \bibfield{editor}{\bibinfo{person}{L.~DARRELL WHITLEY}} (Ed.). \bibinfo{series}{Foundations of Genetic Algorithms}, Vol.~\bibinfo{volume}{2}. \bibinfo{publisher}{Elsevier}, \bibinfo{pages}{153--165}.
\newblock
\showISSN{1081-6593}
\urldef\tempurl%
\url{https://doi.org/10.1016/B978-0-08-094832-4.50016-7}
\showDOI{\tempurl}


\bibitem[Sobania et~al\mbox{.}(2022)]%
        {sobania2022comprehensive}
\bibfield{author}{\bibinfo{person}{Dominik Sobania}, \bibinfo{person}{Dirk Schweim}, {and} \bibinfo{person}{Franz Rothlauf}.} \bibinfo{year}{2022}\natexlab{}.
\newblock \showarticletitle{A comprehensive survey on program synthesis with evolutionary algorithms}.
\newblock \bibinfo{journal}{\emph{IEEE Transactions on Evolutionary Computation}} (\bibinfo{year}{2022}).
\newblock


\bibitem[Spector(2001)]%
        {spector2:2001:gecco}
\bibfield{author}{\bibinfo{person}{Lee Spector}.} \bibinfo{year}{2001}\natexlab{}.
\newblock \showarticletitle{Autoconstructive Evolution: {Push, PushGP, and Pushpop}}. In \bibinfo{booktitle}{\emph{Proceedings of the Genetic and Evolutionary Computation Conference (GECCO-2001)}}, \bibfield{editor}{\bibinfo{person}{Lee Spector}, \bibinfo{person}{Erik~D. Goodman}, \bibinfo{person}{Annie Wu}, \bibinfo{person}{W.~B. Langdon}, \bibinfo{person}{Hans-Michael Voigt}, \bibinfo{person}{Mitsuo Gen}, \bibinfo{person}{Sandip Sen}, \bibinfo{person}{Marco Dorigo}, \bibinfo{person}{Shahram Pezeshk}, \bibinfo{person}{Max~H. Garzon}, {and} \bibinfo{person}{Edmund Burke}} (Eds.). \bibinfo{publisher}{Morgan Kaufmann}, \bibinfo{address}{San Francisco, California, USA}, \bibinfo{pages}{137--146}.
\newblock
\showISBNx{1-55860-774-9}
\urldef\tempurl%
\url{http://hampshire.edu/lspector/pubs/ace.pdf}
\showURL{%
\tempurl}


\bibitem[Spector(2012)]%
        {assessment_spector_2012}
\bibfield{author}{\bibinfo{person}{Lee Spector}.} \bibinfo{year}{2012}\natexlab{}.
\newblock \showarticletitle{Assessment of Problem Modality by Differential Performance of Lexicase Selection in Genetic Programming: A Preliminary Report}. In \bibinfo{booktitle}{\emph{Proceedings of the 14th Annual Conference Companion on Genetic and Evolutionary Computation}} (Philadelphia, Pennsylvania, USA) \emph{(\bibinfo{series}{GECCO '12})}. \bibinfo{publisher}{Association for Computing Machinery}, \bibinfo{address}{New York, NY, USA}, \bibinfo{pages}{401–408}.
\newblock
\showISBNx{9781450311786}
\urldef\tempurl%
\url{https://doi.org/10.1145/2330784.2330846}
\showDOI{\tempurl}


\bibitem[Spector et~al\mbox{.}(2005)]%
        {spector_push3_2005}
\bibfield{author}{\bibinfo{person}{Lee Spector}, \bibinfo{person}{Jon Klein}, {and} \bibinfo{person}{Maarten Keijzer}.} \bibinfo{year}{2005}\natexlab{}.
\newblock \showarticletitle{The Push3 Execution Stack and the Evolution of Control}. In \bibinfo{booktitle}{\emph{Proceedings of the 7th Annual Conference on Genetic and Evolutionary Computation}} (Washington DC, USA) \emph{(\bibinfo{series}{GECCO '05})}. \bibinfo{publisher}{Association for Computing Machinery}, \bibinfo{address}{New York, NY, USA}, \bibinfo{pages}{1689–1696}.
\newblock
\showISBNx{1595930108}
\urldef\tempurl%
\url{https://doi.org/10.1145/1068009.1068292}
\showDOI{\tempurl}


\bibitem[Spector and Robinson(2002)]%
        {spector_genetic_2002}
\bibfield{author}{\bibinfo{person}{Lee Spector} {and} \bibinfo{person}{Alan Robinson}.} \bibinfo{year}{2002}\natexlab{}.
\newblock \showarticletitle{Genetic {Programming} and {Autoconstructive} {Evolution} with the {Push} {Programming} {Language}}.
\newblock \bibinfo{journal}{\emph{Genetic Programming and Evolvable Machines}} \bibinfo{volume}{3}, \bibinfo{number}{1} (\bibinfo{date}{March} \bibinfo{year}{2002}), \bibinfo{pages}{7--40}.
\newblock
\showISSN{1573-7632}
\urldef\tempurl%
\url{https://doi.org/10.1023/A:1014538503543}
\showDOI{\tempurl}


\bibitem[Stanton and Moore(2022)]%
        {stanton_lexicase_2022}
\bibfield{author}{\bibinfo{person}{Adam Stanton} {and} \bibinfo{person}{Jared~M. Moore}.} \bibinfo{year}{2022}\natexlab{}.
\newblock \showarticletitle{Lexicase {Selection} for {Multi}-{Task} {Evolutionary} {Robotics}}.
\newblock \bibinfo{journal}{\emph{Artificial Life}} \bibinfo{volume}{28}, \bibinfo{number}{4} (\bibinfo{date}{Nov.} \bibinfo{year}{2022}), \bibinfo{pages}{479--498}.
\newblock
\showISSN{1064-5462}
\urldef\tempurl%
\url{https://doi.org/10.1162/artl_a_00374}
\showDOI{\tempurl}
\newblock
\shownote{\_eprint: https://direct.mit.edu/artl/article-pdf/28/4/479/2043352/artl\_a\_00374.pdf}.


\bibitem[Yan et~al\mbox{.}(2019)]%
        {YanFPS}
\bibfield{author}{\bibinfo{person}{Xuyang Yan}, \bibinfo{person}{Mohammad Razeghi-Jahromi}, \bibinfo{person}{Abdollah Homaifar}, \bibinfo{person}{Berat~A. Erol}, \bibinfo{person}{Abenezer Girma}, {and} \bibinfo{person}{Edward Tunstel}.} \bibinfo{year}{2019}\natexlab{}.
\newblock \showarticletitle{A Novel Streaming Data Clustering Algorithm Based on Fitness Proportionate Sharing}.
\newblock \bibinfo{journal}{\emph{IEEE Access}}  \bibinfo{volume}{7} (\bibinfo{year}{2019}), \bibinfo{pages}{184985--185000}.
\newblock
\urldef\tempurl%
\url{https://doi.org/10.1109/ACCESS.2019.2922162}
\showDOI{\tempurl}


\bibitem[Zhong et~al\mbox{.}(2005)]%
        {Zhong_tourn_2005}
\bibfield{author}{\bibinfo{person}{Jinghui Zhong}, \bibinfo{person}{Xiaomin Hu}, \bibinfo{person}{Jun Zhang}, {and} \bibinfo{person}{Min Gu}.} \bibinfo{year}{2005}\natexlab{}.
\newblock \showarticletitle{Comparison of Performance between Different Selection Strategies on Simple Genetic Algorithms.}, Vol.~\bibinfo{volume}{2}. \bibinfo{pages}{1115--1121}.
\newblock
\urldef\tempurl%
\url{https://doi.org/10.1109/CIMCA.2005.1631619}
\showDOI{\tempurl}


\end{thebibliography}

\end{document}